\pdfoutput=1
\documentclass[lettersize,journal]{IEEEtran}
\usepackage{spconf,amsmath,graphicx}

\usepackage{algorithm}
\usepackage{algpseudocode}
\usepackage{amsmath}
\usepackage{subfigure}
\usepackage{graphicx}  

\usepackage{graphicx}
\usepackage{amsmath}
\usepackage{amssymb}
\usepackage{booktabs}
\usepackage{wrapfig}
\usepackage{xspace}

\usepackage{graphicx}
\usepackage{amsmath}
\usepackage{amssymb}
\usepackage{booktabs}
\usepackage[marginal]{footmisc}

\usepackage{amsmath, amssymb} 
\usepackage{booktabs}
\usepackage{bm}
\usepackage{textcomp}
\usepackage{gensymb}
\usepackage{pifont}
\usepackage[T1]{fontenc}
\usepackage{multirow}
\usepackage[rightcaption]{sidecap}
\usepackage{makecell}
\usepackage{xspace}
\usepackage{enumitem}
\usepackage{url}
\usepackage{enumerate}
\usepackage{wrapfig}
\usepackage{algorithm}
\usepackage{wrapfig,lipsum,booktabs}
\usepackage{amsmath,amssymb}
\usepackage{xcolor} 

\usepackage[colorlinks=true,linkcolor=blue,citecolor=blue,urlcolor=blue]{hyperref}

\usepackage[capitalize]{cleveref}
\crefname{section}{Sec.}{Secs.}
\Crefname{section}{Section}{Sections}
\Crefname{table}{Table}{Tables}
\crefname{table}{Tab.}{Tabs.}

\makeatletter

\DeclareRobustCommand\onedot{\futurelet\@let@token\@onedot}
\def\@onedot{\ifx\@let@token.\else.\null\fi\xspace}
 
\def\ie{\emph{i.e}\onedot}

\def\etal{\emph{et al}\onedot}

\makeatother

\begin{document}

\title{Unsupervised Search for Ethnic Minorities' \\ Medical Segmentation Training Set}

\name{
Yixiao Chen$^{1}$, Yue Yao$^{2 \star}$, Ruining Yang$^{1}$, Md Zakir Hossain$^{3}$, Ashu Gupta$^{3}$, Tom Gedeon$^{3}$
\thanks{${\star}$ Corresponding author} 
\thanks{© 2025 IEEE.  Personal use of this material is permitted.  Permission from IEEE must be obtained for all other uses, in any current or future media, including reprinting/republishing this material for advertising or promotional purposes, creating new collective works, for resale or redistribution to servers or lists, or reuse of any copyrighted component of this work in other works.}
}

\address{}

\maketitle

\begin{abstract}
This article investigates the critical issue of dataset bias in medical imaging, with a particular emphasis on racial disparities caused by uneven population distribution in dataset collection. Our analysis reveals that medical segmentation datasets are significantly biased, primarily influenced by the demographic composition of their collection sites. For instance, Scanning Laser Ophthalmoscopy (SLO) fundus datasets collected in the United States predominantly feature images of White individuals, with minority racial groups underrepresented. This imbalance can result in biased model performance and inequitable clinical outcomes, particularly for minority populations. To address this challenge, we propose a novel training set search strategy aimed at reducing these biases by focusing on underrepresented racial groups. Our approach utilizes existing datasets and employs a simple greedy algorithm to identify source images that closely match the target domain distribution. By selecting training data that aligns more closely with the characteristics of minority populations, our strategy improves the accuracy of medical segmentation models on specific minorities, \ie, Black. Our experimental results demonstrate the effectiveness of this approach in mitigating bias. We also discuss the broader societal implications, highlighting how addressing these disparities can contribute to more equitable healthcare outcomes. Our code is available at \url{https://github.com/yorkeyao/SnP}.
\end{abstract}



%

\vspace{-0.5em}
\section{Introduction}

Artificial intelligence plays a crucial role in various fields\cite{kenton2019bert,dosovitskiy2020image,yang2024ploc}, especially in medicine, where it is increasingly significant in identifying diseases and abnormalities\cite{ma2024segment,liu2021federated}, especially in medical image segmentation tasks. However, a critical issue that is often overlooked is that minority ethnic groups, such as black patients, are severely underrepresented in existing datasets. This data imbalance can bias the performance of medical models in different racial groups. There are differences between races in anatomical characteristics, such as body shape, breast density, disease distribution, and tissue specificity. These characteristics are critical for accurate medical image segmentation~\cite{gichoya2022ai}. When a model is trained primarily 
\begin{figure}[t]
 \begin{minipage}[c]{0.26\textwidth}
    \includegraphics[width=\textwidth]{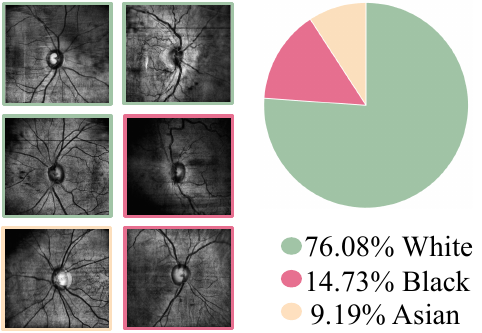}
  \end{minipage}
  \hfill
\begin{minipage}[c]{0.20\textwidth}
    \caption{Image samples and race composition statistics of the SLO fundus dataset. Left columns present the target samples, while the pie chart illustrates the unbalanced distribution across races.
    }
  \label{fig:intro}
  \end{minipage}
  \vspace{-4mm}
\end{figure}
\noindent on data from a single racial group, its accuracy and reliability tend to drop significantly on other racial groups. For example, studies have shown that when predicting the health status of black patients, AI models mistakenly believed that they were healthier than white patients with the same condition, reflecting the injustice of the model in minority groups~\cite{norori2021addressing}.

Traditional medical segmentation models are often biased toward specific races, leading to poor performance among ethnic minority groups. As shown in Fig.~\ref{fig:intro}, the FairSeg dataset is 76\% white patients, with only 24\% black and Asian patients. This imbalance reduces the effectiveness of the model for minorities and may worsen healthcare disparities. To address this, we propose a training set search strategy using a greedy algorithm to better match the data distribution of minority groups, improving model performance by optimizing the selection of images that represent target racial groups, such as black patients.

Specifically, our training set search process consists of three key steps. \textbf{First}, we divide the data pool into K clusters to organize similar data points into structured groups, which provides the basis for subsequent analysis. \textbf{Next}, we calculate the difference in distribution between each clustered group and the target domain (such as the data of black patients) to quantify its similarity. This step provides guidance in selecting the relevant samples. \textbf{Finally}, based on the differences in distribution, we calculate the sampling score and construct an optimized training set by selecting the data points that best match the distribution of the target domain, ensuring that this subset can effectively reduce bias in the dataset and improve the fairness of the model sex. 

Our experimental results show that the greedy search algorithm consistently outperforms random selection in terms of segmentation accuracy and fairness across different training set sizes. The greedy approach leads to significant improvements in both FID scores and Dice/IoU metrics, demonstrating its effectiveness in mitigating dataset bias and enhancing model performance on underrepresented minority groups. This validates the potential of our strategy to reduce racial disparities in medical imaging models. Through this strategy, we are not only able to improve the accuracy of medical image segmentation models in minority groups, but we also hope that this work can play a positive role in reducing racial inequalities in healthcare.

\section{Related Work}

\textbf{Medical Image Segmentation} is critical for tasks like diagnosis and treatment evaluation. Traditional methods like thresholding faced challenges with complex images, whereas deep learning enhanced segmentation by learning features. Meta AI’s SAM \cite{kirillov2023segment} based on the Transformer \cite{vaswani2017attention} used large-scale pre-training to handle diverse tasks. SAMed \cite{zhang2023customized} adapted SAM for medical imaging but lacked racial labels in the training data, reducing accuracy in diverse populations. This limitation is significant in fundus image segmentation, where retinal differences can cause biased results.


\textbf{Medical Fairness Datasets} aimed to mitigate healthcare disparities, especially for minority groups. Despite advancements in disease detection, model bias persisted. Most fairness datasets \cite{irvin2019chexpert, johnson2019mimic, groh2021evaluating} focused on image classification and limited attributes like age and race, overlooking segmentation tasks. The Harvard-FairSeg dataset \cite{tian2024fairseg} addressed this by offering a large segmentation dataset for optic disc and cup in SLO fundus images, annotated for six attributes, supporting fairness research across diverse demographics.

\textbf{Model-Centric Fairness Learning} addressed bias from data imbalance, which affected model predictions. Fairness strategies were applied in three stages: preprocessing (e.g., resampling) reduced bias before training but often lowered efficiency due to data operations \cite{ramaswamy2021fair, zhang2020towards, park2022fair}. Internal processing integrated fairness constraints into training, often by modifying loss functions, which sometimes reduced overall accuracy \cite{ sarhan2020fairness, zafar2017fairness, zhang2018mitigating}. Post-processing applied corrections after prediction to improve fairness but could not fully address training biases \cite{wang2022fairness, kim2019multiaccuracy, lohia2019bias}.

\textbf{Training Set Search} included innovative data-efficient training approaches \cite{yan2020neural, yao2023large, yang2024data}, focusing on selecting representative training data to optimize learning. Xu \etal developed a classifier that used attention-based multi-scale feature extraction to select informative subsets from unlabeled data \cite{xu2019positive}, while Yan \etal focused on pre-training data selection to better align with target domains \cite{yan2020neural, settles2009active}. SnP \cite{yao2023large} dynamically adjusted the training set size based on budget constraints but was specifically designed for re-identification (re-ID) tasks rather than general use.

\begin{figure}[t]
    \centering
    \includegraphics[width=0.48\textwidth]{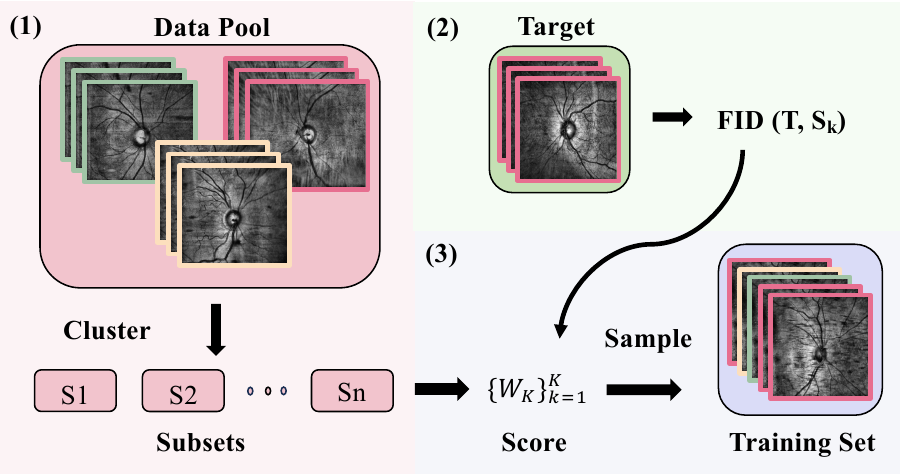}
    \caption{The proposed minority-specific training set search algorithm consists of three steps. First, the data pool is partitioned into K clusters. Second, the distributional differences between the clusters and the target are computed. Third, sampling scores are calculated, and the searched training set is constructed based on these scores.}
    \label{fig:method}
\end{figure}

\section{Method}

\subsection{Motivation}

Dataset bias commonly exists in medical datasets, especially concerning different racial groups. Depending on how medical data are collected and the racial demographics involved, there are four major types of racial bias in existing medical datasets. We provide examples of each type in Fig.~\ref{fig:intro}. This becomes problematic when the training and testing data have different racial distributions. Suppose that we are targeting a medical treatment for a black person, such a biased training dataset will lead to a significant performance. Given a target domain with a specific racial bias, we aim to identify a target-specific training set with a similar distribution or bias to target. To this end, we propose a training set search algorithm for ethnic minorities.

\subsection{The Minority-specific Search Algorithm}
Given a data pool $D$ and a target set $T$, our objective is to sample data from $D$ and construct a searched training set $\hat{P}$ such that $\text{FID}(T, \hat{P})$ is minimized. The Fr\'{e}chet Inception Distance (FID) is defined as:
\begin{align}
\mbox{FID}({\mathbf S}^*, \mathcal{D}_T) =& \left \| \bm{\mu}_s - \bm{\mu}_t  \right \|^{2}_{2} \nonumber +  \\  
& \mathrm{Tr}(\bm{\Sigma}_s + \bm{\Sigma}_t -2 (\bm{\Sigma}_s \bm{\Sigma}_t)^{\frac{1}{2}}).
\label{eq:fid} 
\end{align}

\noindent In Eq. \ref{eq:fid}, $\bm{\mu}_s \in \mathbb{R}^d$ and $\bm{\Sigma}_s \in \mathbb{R}^{d\times d}$ are the mean and covariance matrix of the image descriptors of ${\mathbf S}^*$, respectively. $\bm{\mu}_t$ and $\bm{\Sigma}_t$ are those of $\mathcal{D}_T$. $\mathrm{Tr}(.)$ represents the trace of a square matrix.

Through this approach, we aim to generate a high-quality training set that closely matches the target domain. As illustrated in the Fig.~\ref{fig:method}, our training set search method can be divided into three main steps:

\textbf{First}, we partition the data pool D into K subsets  $\{S_k\}_{k=1}^K$. To do this, we compute the identity-averaged feature for each image, reducing dimensionality while retaining global identity information. Then, we apply the k-means algorithm to cluster these features into  K  groups, forming  K  subsets where each contains images associated with identities within a cluster. This approach efficiently organizes similar data points for further analysis.

\textbf{Next}, we compute the Fréchet Inception Distance (FID) between the target set $T$ and each subset $S_k$, denoted as $\text{FID}(T, S_k)$. The FID measures the similarity between two distributions of data, where lower FID values indicate that the subset $S_k$ is more similar to the target set $T$ in the feature space. By calculating the FID, we quantify the difference between each subset and the target domain, providing guidance for the subsequent sampling process.

\begin{algorithm}[t]
\caption{Greedy Search for Training Set Selection}
\begin{algorithmic}[1]
\State \textbf{Input:} Data pool $D$, Target set $T$, Number of clusters $K$, Number of samples $N$
\State \textbf{Output:} Searched training set $\hat{P}$

\State Cluster $(D, K) \rightarrow \{S^1, \dots, S^K\}$ \Comment{Apply k-means clustering}

\For {each subset $S_k \in \{S_k\}_{k=1}^K$}
    \State Compute $\text{FID}(T, S_k)$ \Comment{Calculate FID for subsets}
\EndFor
\State $w_k = \text{softmax}(-\text{FID}(T, S_k))$

\State Assign probability weight $\frac{w_k}{|S_k|}$ for each identity

\State Sample $N$ examples from $D$ based on the probabilistic weights of the identities
\State Construct the searched training set $\hat{P}$

\State \textbf{Return:} $\hat{P}$

\end{algorithmic}
\end{algorithm}

\textbf{Finally}, we calculate a sampling score $\{w_k\}_{k=1}^K$ for each subset based on the FID values $\{\text{FID}(T, S_k)\}_{k=1}^K$ and assign probabilistic weights to each identity. 
Specifically, the sampling score is computed based on the negative of the FID values, as there is a negative correlation between FID and the quality of the training set. The sampling score is calculated as follows:
\begin{equation}
    \{w_k\}_{k=1}^K = \text{softmax}(\{-\text{FID}(T, S_k)\}_{k=1}^K)
\label{eq:eq2} 
\end{equation}
In Eq. \ref{eq:eq2}, we negate each FID value and apply the softmax function to normalize the values into probabilistic weights. Subsequently, each identity in a cluster is assigned a probabilistic weight of $\frac{w_k}{|S_k|}$, where $|S_k|$ denotes the total number of identities in subset $S_k$. This ensures that each identity is sampled fairly according to its weight and that the sampled training set better aligns with the distribution of the target set. Finally, we sample $N$ examples from the data pool $D$ using the probabilistic weightings of the identities, generating the searched training set.

\subsection{The Minority-specific Segmentation Model}

\begin{figure}[t]
 \begin{minipage}[c]{0.26\textwidth}
    \includegraphics[width=\textwidth]{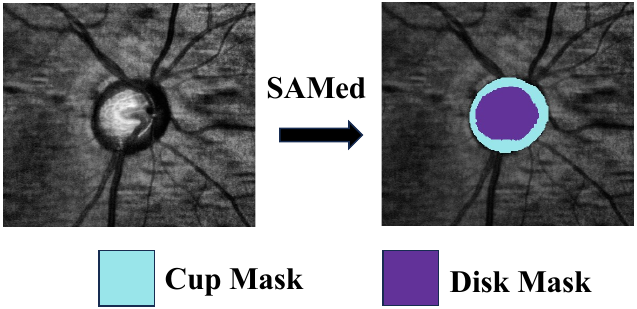}
  \end{minipage}
  \hfill
\begin{minipage}[c]{0.20\textwidth}
    \caption{SLO fundus segmentation task using SAMed~\cite{zhang2023customized}, illustrating the segmentation of the optic cup and disk with corresponding masks applied.
    }
  \label{fig:seg_task}
  \end{minipage}
  \vspace{-4mm}
\end{figure}

Shown in Fig.~\ref{fig:seg_task}, we utilized SAMed \cite{zhang2023customized}, an adaptation of SAM \cite{kirillov2023segment}, designed for medical imaging tasks targeting minority groups. 
We use the SAMed algorithm as to the best of our knowledge it is the SOTA medical segmentation model~\cite{zhang2023customized}. A key component of SAMed is the integration of Low-Rank Adaptation (LoRA) \cite{hu2021lora}, which facilitates efficient fine-tuning of the SAM image encoder. This method reduces the need for full retraining while maintaining performance, especially in contexts where labeled medical data for minority groups is limited. LoRA modules are incorporated into both the prompt encoder and mask decoder, optimizing segmentation accuracy with minimal computational overhead. By fine-tuning SAMed on minority-specific datasets, the model enhances its ability to identify complex structures and rare pathologies, maintaining robust performance despite data imbalances.

\section{Experimental}

\subsection{Settings}

\textbf{DataSet.} The FairSeg dataset consists of 10,000 fundus images collected via scanning laser ophthalmoscopy (SLO) from 10,000 individuals between 2010 and 2021, including 919 Asian, 1,473 Black, and 7,608 White participants. It provides segmentation labels for the optic disc and cup. The test-set contains 2,000 samples with 169 Asian, 299 Black, and 1,532 White participants. In our experiments, we maintained the distribution of the training and test sets, conducting the search on the training set while testing on the dataset samples.

\textbf{Training settings.} 
In the training process, we employed a composite loss function, combining cross-entropy loss and Dice loss, similar to the approach used by Yu \etal. The balance between these two loss components was adjusted using weight parameters $\lambda$1 and $\lambda$2, respectively. For optimization, we selected the AdamW optimizer and applied an exponential learning rate decay strategy, setting the initial learning rate to 0.005. The momentum, weight decay, and batch size were set to 0.9, 0.1, and 28, respectively.

\textbf{Search configuration.} 
The sample sets we selected were much smaller than the original dataset and test set. Running tests for many epochs risked overfitting and could affect the validity of our results. To address this, we applied early stopping for the 100, 500, and 1000-sample sets selected through two sampling strategies. Specifically, we evaluated the model using weights from the 10th, 5th, and 3rd epochs, respectively, for a more reliable performance comparison.

\textbf{Computation Resources.} 
We run experiments on an 8-GPU machine. The 8-GPU machine has 8 NVIDIA 3090 GPUs and its CPU is AMD EPYC 7343 Processor. We use PyTorch version 1.12 for the training set search for segmentation tasks.

\begin{table}
\centering
\footnotesize
\setlength{\tabcolsep}{1.3mm}
\caption{Performance Comparison of Training Set Selection}
\label{tab:tss}
\begin{tabular}{c|c|c|cc|cc|cc} 
\hline\hline
\multicolumn{2}{c|}{\multirow{2}{*}{training set}} & \multirow{2}{*}{FID} & \multicolumn{2}{c|}{Overall}                        & \multicolumn{2}{c|}{Cup}                            & \multicolumn{2}{c}{Rim}                             \\ 
\cline{4-9}
\multicolumn{2}{c|}{}                              &                       & \multicolumn{1}{l}{Dice} & \multicolumn{1}{l|}{IoU} & \multicolumn{1}{l}{Dice} & \multicolumn{1}{l|}{IoU} & \multicolumn{1}{l}{Dice} & \multicolumn{1}{l}{IoU}  \\ 
\hline
\multirow{2}{*}{100} & random                 & 170.80                & 49.15                    & 36.84                    & 64.71                    & 51.15                    & 33.59                    & 21.54                    \\
                          & greedy                 & \textbf{99.51}        & \textbf{50.76}           & \textbf{37.68}           & \textbf{65.14}           & \textbf{52.00}           & \textbf{36.39}           & \textbf{23.35}           \\ 
\hline
\multirow{2}{*}{500}  & random                 & 130.36                & 64.02                    & 50.75                    & 68.61                    & 57.53                    & 59.43                    & 43.97                    \\
                          & greedy                 & \textbf{75.38}        & \textbf{65.59}           & \textbf{52.14}           & \textbf{69.07}           & \textbf{57.58}           & \textbf{62.11}           & \textbf{46.49}           \\ 
\hline
\multirow{2}{*}{1000} & random                 & 139.50                & 72.37                    & 59.69                    & 79.39                    & 68.66                    & 65.35                    & 50.72                    \\
                          & greedy                 & \textbf{94.34}        & \textbf{74.75}           & \textbf{62.09}           & \textbf{79.54}           & \textbf{68.76}           & \textbf{69.96}           & \textbf{55.42}           \\
\hline\hline
\end{tabular}
\end{table}

\subsection{Results}

\textbf{Random sampling} \textbf{\textit{vs.}} \textbf{Greedy sampling.} 
We applied our search algorithm using samples from the black population as the target set, and the results in Table~\ref{tab:tss} compare segmentation accuracy across different search methods and selection sizes. Segmentation performance is evaluated using the Dice coefficient and Intersection over Union (IoU).

The first column shows that the subset selected by the greedy method is more similar to the target set in terms of feature-level similarity than the randomly selected subset. Moreover, comparisons across various selection sizes indicate that the greedy-selected subset consistently outperforms the random selection in both performance metrics. This implies that subsets more aligned with the target domain tend to achieve faster convergence during training.
Overall, these results suggest that our unsupervised method helps reduce data bias, improving model fairness after training.

\textbf{Computational cost in searching for a training set.}
As illustrated in Fig.~\ref{fig:method}, our training set search process involves three distinct steps. Using 8000 samples as the source and 300 samples as the target, feature extraction and clustering take approximately 300 seconds. Subsequently, computing the FID takes around 600 seconds. The time required for image sampling is negligible, resulting in a total run time of approximately 900 seconds for our algorithm. To be mentioned, the results of feature extraction and clustering can be reused, and changing the sample rate will not lead to additional cost in the first step.

\begin{figure}[t]
 \begin{minipage}[c]{0.26\textwidth}
    \includegraphics[width=\textwidth]{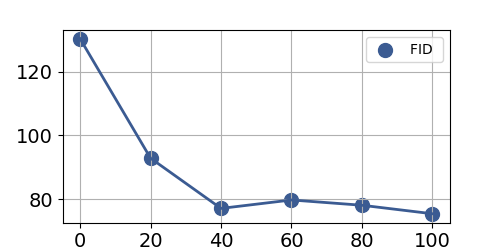}
  \end{minipage}
  \hfill
\begin{minipage}[c]{0.20\textwidth}
    \caption{Impact of the number of clusters K to the domain gap between searched and target.
    }
  \label{fig:para_study}
  \end{minipage}
  \vspace{-4mm}
\end{figure}

\textbf{Numbers of clusters K and images N of searched training set Set}. In our approach, we partition the data pool into K clusters using image features and sample N identities to construct the searched training set. Through experimentation shown in Fig.~\ref{fig:para_study}, we observed that selecting an excessively small or large value for K can slightly degrade the quality of the searched training set. Moreover, as N increases, the results tend to stabilize. Therefore, we set K to an intermediate value of 100 and N from (100, 500, 1000) to achieve relatively optimal results.

\section{Discussion}

\textbf{Broad social impact}. Accurately representing underrepresented minority populations, like black individuals, in medical datasets is vital for ensuring equitable healthcare outcomes. Biases in segmentation models can lead to misdiagnosis or inadequate treatment, reinforcing existing health disparities. By adopting strategies to reduce these biases, such as the proposed training set search approach, we enhance model performance and promote fairer diagnostic practices. This helps reduce systemic healthcare inequalities, providing minority groups with better access to accurate diagnoses and personalized treatments. As AI’s role in healthcare expands, fairness in medical models is essential for fostering trust and inclusivity.

\textbf{Further improvement}. Our experiments used InceptionV3 pre-trained on ImageNet, a strong model for image classification and feature extraction in natural images. However, medical images, such as fundus data, may benefit from more specialized networks. In future work, replacing InceptionV3 or incorporating multi-scale feature fusion strategies could capture finer image details, improving clustering and ranking accuracy. This would narrow the gap between training and target sets, leading to even better results.

\section{Conclusion}

This paper addresses racial bias in medical imaging datasets, particularly in segmentation tasks. Our proposed training set search strategy effectively targets underrepresented racial groups by selecting images that match the target domain’s distribution, offering a promising approach to mitigate bias. Our method does not rely on labels, making it applicable not only to segmentation tasks but also potentially to other medical imaging applications.

However, the method is limited by the diversity of available datasets and race labeling, and future work should focus on integrating more diverse data sources and addressing other forms of bias. Our study highlights the need for fairness in medical AI systems, as biased models can negatively impact healthcare outcomes for minority populations.

\newpage
\bibliographystyle{IEEEbib}
\bibliography{references}

\end{document}